\PassOptionsToPackage{unicode}{hyperref}
\PassOptionsToPackage{hyphens}{url}
\documentclass[
]{article}
\usepackage{amsmath,amssymb}
\usepackage{lmodern}
\usepackage{ifxetex,ifluatex}
\ifnum 0\ifxetex 1\fi\ifluatex 1\fi=0 
  \usepackage[T1]{fontenc}
  \usepackage[utf8]{inputenc}
  \usepackage{textcomp} 
\else 
  \usepackage{unicode-math}
  \defaultfontfeatures{Scale=MatchLowercase}
  \defaultfontfeatures[\rmfamily]{Ligatures=TeX,Scale=1}
\fi
\IfFileExists{upquote.sty}{\usepackage{upquote}}{}
\IfFileExists{microtype.sty}{
  \usepackage[]{microtype}
  \UseMicrotypeSet[protrusion]{basicmath} 
}{}
\makeatletter
\@ifundefined{KOMAClassName}{
  \IfFileExists{parskip.sty}{%
    \usepackage{parskip}
  }{
    \setlength{\parindent}{0pt}
    \setlength{\parskip}{6pt plus 2pt minus 1pt}}
}{
  \KOMAoptions{parskip=half}}
\makeatother
\usepackage{xcolor}
\IfFileExists{xurl.sty}{\usepackage{xurl}}{} 
\IfFileExists{bookmark.sty}{\usepackage{bookmark}}{\usepackage{hyperref}}
\hypersetup{
  pdftitle={Transforming Feature Space to Interpret Machine Learning Models},
  pdfauthor={Alexander Brenning},
  pdfkeywords={interpretable machine learning, dataset-level post-hoc interpretation, predictive modelling, model visualization, feature space transformation},
  hidelinks,
  pdfcreator={LaTeX via pandoc}}
\usepackage[margin=1in]{geometry}
\usepackage{longtable,booktabs,array}
\usepackage{calc} 
\usepackage{etoolbox}
\makeatletter
\patchcmd\longtable{\par}{\if@noskipsec\mbox{}\fi\par}{}{}
\makeatother
\IfFileExists{footnotehyper.sty}{\usepackage{footnotehyper}}{\usepackage{footnote}}
\makesavenoteenv{longtable}
\usepackage{graphicx}
\makeatletter
\def\maxwidth{\ifdim\Gin@nat@width>\linewidth\linewidth\else\Gin@nat@width\fi}
\def\maxheight{\ifdim\Gin@nat@height>\textheight\textheight\else\Gin@nat@height\fi}
\makeatother
\setkeys{Gin}{width=\maxwidth,height=\maxheight,keepaspectratio}
\makeatletter
\def\fps@figure{htbp}
\makeatother
\providecommand{\tightlist}{%
  \setlength{\itemsep}{0pt}\setlength{\parskip}{0pt}}
\ifluatex
  \usepackage{selnolig}  
\fi
\newlength{\cslhangindent}
\newlength{\csllabelwidth}
\newenvironment{CSLReferences}[2] 
 {
  \setlength{\parindent}{0pt}
  \ifodd #1 \everypar{\setlength{\hangindent}{\cslhangindent}}\ignorespaces\fi
  \ifnum #2 > 0
  \setlength{\parskip}{#2\baselineskip}
  \fi
 }%
 {}
\usepackage{calc}

\title{Transforming Feature Space to Interpret Machine Learning Models}
\author{Alexander Brenning}
\date{Michael Stifel Center Jena for Data-Driven and Simulation Science (MSCJ), and Geographic Information Science group, Department of Geography, Friedrich Schiller University Jena, Germany}

\begin{document}
\maketitle
\begin{abstract}
Model-agnostic tools for interpreting machine-learning models struggle to summarize the joint effects of strongly dependent features in high-dimensional feature spaces, which play an important role in pattern recognition, for example in remote sensing of landcover. This contribution proposes a novel approach that interprets machine-learning models through the lens of feature space transformations. It can be used to enhance unconditional as well as conditional post-hoc diagnostic tools including partial dependence plots, accumulated local effects plots, or permutation feature importance assessments. While the approach can also be applied to nonlinear transformations, we focus on linear ones, including principal component analysis (PCA) and a partial orthogonalization technique. Structured PCA and diagnostics along paths offer opportunities for representing domain knowledge. The new approach is implemented in the R package \texttt{wiml}, which can be combined with existing explainable machine-learning packages. A case study on remote-sensing landcover classification with 46 features is used to demonstrate the potential of the proposed approach for model interpretation by domain experts.
\end{abstract}

\hypertarget{introduction}{%
\section{Introduction}\label{introduction}}

Interpreting complex nonlinear machine-learning models is an inherently difficult task. A common approach is the post-hoc analysis of black-box models for dataset-level interpretation (Murdoch et al. 2019) using model-agnostic techniques such as the permutation-based variable importance, and graphical displays such as partial dependence plots that visualize main effects while integrating over the remaining dimensions (Molnar, Casalicchio, and Bischl 2020).

These tools are so far limited to displaying the relationship between the response and one (or sometimes two) predictor(s), while attempting to control for the influence of the other predictors. This can be rather unsatisfactory when dealing with a large number of highly correlated predictors, which are often semantically grouped. While the literature on explainable machine learning has often focused on dealing with dependencies affecting \emph{individual} features, e.g.~by introducing conditional diagnostics (Strobl et al. 2008; Molnar, König, Bischl, et al. 2020), no practical solutions are available yet for dealing with model interpretation in high-dimensional feature spaces with strongly dependent features (Molnar, Casalicchio, and Bischl 2020; Molnar, König, Herbinger, et al. 2020).

These situations routinely occur in environmental remote sensing and other geographical and ecological analyses (Landgrebe 2002; Zortea, Haertel, and Clarke 2007), which motivated the present proposal to enhance existing model interpretation tools by offering a new, transformed perspective. For example, vegetation `greenness' as a measure of photosynthetic activity is often used to classify landcover or land use from satellite imagery acquired at multiple time points throughout the growing season (Peña and Brenning 2015; Peña, Liao, and Brenning 2017). Spectral reflectances of equivalent spectral bands (the features) are usually strongly correlated within the same phenological stage since vegetation characteristics vary gradually. Similarly. when using texture features to characterize image structure based on a filter bank, features with similar filter settings can be strongly correlated, as is in the case in our case study (Brenning, Long, and Fieguth 2012).

Although it may be tempting in these situations to use feature engineering or feature selection techniques to reduce the complexity of feature space, experience shows that this may lead to a decline in predictive performance. Also, re-training a model using modified features is not normally an option in post-hoc analyses of machine-learning models.

Considering these challenges, and the inherent need to reduce the complexity at the time of interpreting an already trained model, a novel strategy to visualize machine-learning models along cross-section through feature space is proposed in this paper. In many situations, principal components (PCs) offer a particularly appealing perspective onto feature space from a practitioner's point of view, although the proposed approach is not limited to this transformation. In addition, a modification is proposed that focuses on subgroups of features and their principal axes in order to allow for a more structured approach to model interpretation that more consistent with the data analyst's domain knowledge.

Turning our attention back to the importance of individual features, an orthogonalization technique that can be used to single out the effect of individual features on model predictions, avoiding the sometimes complex structure of PCs. This approach can, in principle, be applied to arbitrary paths through feature space, such as nonlinear curves defined by domain-specific perspectives, or data-driven transitions between clusters of observations.

The proposed approaches can be combined with commonly used plot types and diagnostics including partial dependence plots, accumulated local effects (ALE) plots, and permutation-based variable importance measures, among other model-agnostic techniques that only have access to the trained model (Apley and Zhu 2020; Molnar 2019). While the focus of this contribution is on visualizing main effects, analyses of conditional relationships may also benefit from this perspective (Strobl et al. 2008; Molnar, König, Bischl, et al. 2020).

\hypertarget{proposed-method}{%
\section{Proposed Method}\label{proposed-method}}

Let's consider a regression or classification model
\[\hat{f}:\mathbf{x}\mapsto\hat{f}(\mathbf{x})\in\mathbb{R}\]
that was fitted to a training sample \(L\) in the (original, untransformed) \(p\)-dimensional feature space \(X\subset\mathbb{R}^p\). I will assume \(\hat{f}(\mathbf{x})\in\mathbb{R}\); in the case of classification problems, \(\hat{f}(\mathbf{x})\) shall therefore represent predictions of some real-valued quantity such as the probability or logit of a selected target class. One of the features, referred to as \(x_s\), is selected as the feature of interest, and the remaining features are denoted by \(\mathbf{x}_C\).

\hypertarget{example-partial-dependence-plots}{%
\subsection{Example: partial dependence plots}\label{example-partial-dependence-plots}}

In this situation, the partial dependence plot of \(\hat{f}\) with respect to \(x_S\) can formally be defined as
\begin{align*}
\hat{f}_{x_S,PDP}(x_S) &= E_{\mathbf{X}_C}\!\left[\hat{f}(x_S,\mathbf{X}_C)\right]\\
&= \int_{\mathbf{x}_C}\!\hat{f}(x_S,\mathbf{x}_C)\,d{}P(\mathbf{x}_C)
\end{align*}
(Molnar 2019). This plot, which can be generalized to more than one \(x_s\) dimension, was introduced by Friedman (2001) to visualize main effects of predictors in machine-learning models.

Partial dependence plots have some disadvantages such as the extrapolation of \(\hat{f}\) beyond the region in \(X\) for which training data is available (Apley and Zhu 2020; Molnar, König, Herbinger, et al. 2020). This is especially the case when predictors are strongly correlated, as in our case study. Nevertheless, without loss of generality, this simple plot type helps to illustrate the proposed approach.

\hypertarget{transformed-feature-space}{%
\subsection{Transformed feature space}\label{transformed-feature-space}}

When several predictors are strongly correlated and/or express the same domain-specific concept such as `early-season vegetation vigour' in vegetation remote-sensing, we may be more interested in exploring the overall effect of these predictors. Principal component analysis (PCA) and related data transformation techniques such as factor analysis are tools that are often used by practitioners to synthesize and interpret multivariate data Basille et al. (2008).

More generally speaking, we could think of bijective (invertible) transformation function
\[
\mathbf{T}: X \rightarrow W\subset\mathbb{R}^p,\quad \mathbf{w} = \mathbf{T}(\mathbf{x})
\]
that can be used to re-express the features in our data set. We will assume that \(\mathbf{T}\) is continuous and differentiable. PCA is one such example.

Through the composition of the back transformation \(\mathbf{T}^{-1}\) and the model function \(\hat{f}\), we can now formally define a model \(\hat{g}\) on \(W\),
\[
\hat{g} := \hat{f}\circ\mathbf{T}^{-1}
\]
which predicts the real-valued response based on `data' in \(W\) although it was trained using a learning sample \(L\subset X\) in the untransformed space.

We can use this to formally re-express the partial dependence plot as a function of \(w_s\):
\begin{align*}
\hat{f}_{w_S,PDP}(w_S) &= E_{w_C}\!\left[(\hat{f}\circ \mathbf{T}^{-1})(w_S,w_C)\right]\\&=\int_{w_C}\!(\hat{f}\circ \mathbf{T}^{-1})(w_S,w_C)\,d{}P\mathbf{w}_C
\end{align*}

Note that \(\mathbf{T}^{-1}\), when used only on data in \(Im_\mathbf{T}(X) := \mathbf{T}(X)\), does not create \(\mathbf{x}\) values outside the data-supported region \(X\), and it therefore avoids extrapolation of \(\hat{f}\).

Also, when choosing PCA for \(\mathbf{T}\) the \(\mathbf{w}\) variables in \(\mathbf{T}(L)\) are linearly independent, and statistically independent if \(L\) arises from a multivariate normal distribution. Thus, the PCA approach overcomes one of the limitations of partial dependence plots and broadens their applicability.

\hypertarget{orthogonalization-approaches}{%
\subsection{Orthogonalization approaches}\label{orthogonalization-approaches}}

In some instances, PCs (and other multivariate transformations) of large and complex feature sets can be difficult to interpret, and analysts would therefore like to focus on individual features that are perhaps `representative' of a larger group of features - for example, vegetation greenness in mid-June may be a good proxy for vegetation greenness a few weeks earlier and later, as expressed by other features in the feature set (Peña and Brenning 2015).

This can be addressed by proposing a transformation of \(X\) in which \(w_s := x_s\) is retained, while making the remaining base vectors linearly independent of \(x_s\). This can be achieved through \emph{orthogonalization}:

\[
w_i := x_i - b_i x_s
\]
where \(b_i\) equals Pearson's correlation coefficient of \(x_s\) and \(x_i\), and where we assume for simplicity of notation that all features are zero-mean with a unit standard deviation.

This defines a linear transformation \(\mathbf{T}:X\rightarrow W\), which can be represented by its coefficient matrix. Note that \(\mathbf{T}\) can be inverted using
\[
x_i = w_i + b_i w_s
\]
since \(x_s = w_s\), and assuming that all \(b_i < 1\). A related iterative orthogonalization approach has previously been proposed in the context of feature ranking (Adebayo and Kagal 2016).

\hypertarget{following-paths}{%
\subsection{Dependence plots along paths}\label{following-paths}}

Data analysts may more generally want to visualize the effect of a real-valued function of multiple features. As an example, knowing that several features are strongly correlated, how does the response vary with the mean value of these features, or more generally a linear combination? This information is sometimes hidden in an ocean of individual main effects plots or variable importance measures.

In other situations, there may be simple process-based models that have the potential to provide deeper insights into black-box models. These models may be candidates for an enhancement of feature space, or they might express specific theories or hypotheses.

Any of these transformations can be thought of as a function
\[
\mathbf{p}:P\subset\mathbb{R}\rightarrow X\subset\mathbb{R}^p,\quad t\mapsto \mathbf{p}(t)
\]
that defines a one-dimensional path in feature space.

In different use cases there may be different ways of constructing paths of interest to the data analyst:

\begin{itemize}
\tightlist
\item
  A group of strongly positively correlated features could be averaged to obtain an overall signal, or contrasts between groups of features could be calculated.
\item
  A linear path can be drawn from one cluster centre to another, where cluster centres \(\mathbf{c}_1, \ldots, \mathbf{c}_k\in X\) are obtained by unsupervised clustering in feature space (e.g., \(k\)-means). The path between clusters \(1\) and \(2\) is simply defined as \(\mathbf{c}_1+t\mathbf{c}_2\), etc.
\item
  A linear path between user-defined points in feature space, e.g.~in remote sensing so-called endmembers representing spectral characteristics of `pure' surface types such as asphalt or water (Somers et al., 2016).
\end{itemize}

Evidently, creating a dependence plot for a feature \(x_s\) is just a special case that follows one of the base vectors of feature space. In other words, without loss of generality we can formally include \(t\) as an additional feature in our feature set and denote it with \(x_s\). This is just a formal inclusion, without actually offering this new variable to the model for training.

For a formally more accurate treatment, and keeping \(t\) out of the feature space \(X\subset\mathbb{R}^p\), we write
\[
w_s := t
\]
and construct the \(w_i\)'s as in the previous section by orthogonalization, however this time for all \(i=1, \ldots, p\). Thus, \(w_s\) comes on top of the other \(p\) dimensions, and thus \(W\subset\mathbb{R}^{(p+1)}\).

Due to the redundancy of \(w_s\), the transformation function can now be defined as a mapping \(\mathbf{T}: W\rightarrow X\) from \((p+1)\)- to \(p\)-dimensional space, with the \(x_i\)'s being recovered from the \(w_i\)'s as in the previous section.

\hypertarget{other-model-agnostic-plots}{%
\subsection{Other model-agnostic plots}\label{other-model-agnostic-plots}}

The same principles outlined in the previous section can be applied to ALE plots and related model-agnostic tools, including permutation-based variable importance and their conditional modifications (see reviews by Molnar et al., 2020a and Molnar, 2021). Also, this is not limited to \(x_s\in\mathbb{R}\) - this principle equally applies to bivariate \(\mathbf{x}_s\in\mathbb{R}^2\) relationships, which can be used to display pairwise interactions. Clearly, in a high-dimensional situation, the need to reduce dimensionality in post-hoc model interpretation is even more pressing when interpreting up to \(p*(p-1)/2\) pairwise interactions, and the proposed approach offers a practical tool to address this in situations where dimension reduction is viable.

\hypertarget{implementation}{%
\section{Implementation}\label{implementation}}

The proposed methods are provided as an R package \texttt{wiml} (\url{https://github.com/alexanderbrenning/wiml}). It implements transformation functions called `warpers' based on PCA (of all features or a subset of features), structured PCA (for multiple groups of features), and feature orthogonalization, all of which are based on rotation matrices and therefore share a common core. Due to the modular and object-oriented structure, users can easily implement their own transformations without requiring changes to the package.

These warpers can be used to implement the composition \(\hat{f}\circ\mathbf{T}^{-1}\) by `warping' a fitted machine-learning model. The resulting object behaves like a regular fitted machine-learning model in R, offering an adapted \texttt{predict} method. From a user's perspective, the resulting model feels like it had been fitted to the transformed data \(\mathbf{T}^{-1}(L)\), except that it hasn't. This `warped' fitted model can, in principle, be used with any model-agnostic tool that doesn't require refitting. An implementation of the composition \(f\circ\mathbf{T}^{-1}\) involving the untrained model \(f\) is also available; this can be used for drop and re-learn or permute and re-learn techniques (Hooker and Mentch 2019).

The package has been tested and works well with the \texttt{iml} package for interpretable machine learning (Molnar, Bischl, and Casalicchio 2018), but it can also be combined with other frameworks since it only builds thin wrappers around standard R model functions. Initial tests with the \texttt{DALEX} framework for explainable machine-learning (Biecek 2018) and its interactive environment \texttt{modelStudio} (Baniecki and Biecek 2019) have been successful, as have been tests with the \texttt{pdp} package (Greenwell 2017).

\hypertarget{case-study}{%
\section{Case Study}\label{case-study}}

The potential of the proposed methods will be demonstrated in a case study from land cover classification, which is a common machine-learning task in environmental remote sensing (e.g., Mountrakis, Im, and Ogole 2011; Peña and Brenning 2015). One particularly challenging task is the detection of rock glaciers, which, unlike `true' glaciers, do not present visible ice on their surface; they are rather the visible expression of creeping ice-rich mountain permafrost. In the present case study, we look at a subset of a well-documented data set consisting of a sample of 1000 labelled point locations (500 presence and 500 absence locations of flow structures on rock glaciers) in the Andes of central Chile (Brenning, Long, and Fieguth 2012).

There are 46 features in total, which are divided into two unequal subsets: Six features are terrain attributes (local slope angle, potential incoming solar radiation, mean slope angle of the catchment area, and logarithm of catchment height and catchment area), which are proxies for processes related to rock glacier formation. The other 40 features are Gabor texture features (Clausi and Jernigan 2000), which are designed to detect the furrow-and-ridge structure in high-resolution (\(1\ \textrm{m}\times 1\ \textrm{m}\)) satellite imagery, in this case panchromatic IKONOS imagery (see Brenning, Long, and Fieguth 2012 for details). The 40 Gabor features correspond to different filter bandwidths (5, 10, 20, 30 and 50 m), anisotropy factors (1 or 2), and types of aggregation over different filter orientations (minimum, median, maximum, and range).

Texture features with similar filter settings are often strongly correlated with each other. This is especially true for minimum and median aggregation with otherwise equal settings, and for maximum and range aggregation. Overall, the median of each feature's strongest Pearson correlation is 0.92 (minimum: 0.80). Correlations among terrain attributes are much smaller (median strongest correlation: 0.60). Terrain attributes and texture features are weakly correlated (maximum correlation: 0.32). Correlation statistics are very similar for Spearman's rank-based correlation.

\begin{longtable}[]{@{}llll@{}}
\caption{\label{tab:accuracies}Overall accuracies of random forest classification of rock glaciers using the entire feature set, and omitting either the terrain attributes or the texture features.}\tabularnewline
\toprule
& All features & Only terrain attributes & Only texture features \\
\midrule
\endfirsthead
\toprule
& All features & Only terrain attributes & Only texture features \\
\midrule
\endhead
Accuracy & 0.801 & 0.759 & 0.696 \\
Decrease in accuracy & 0.000 & 0.042 & 0.105 \\
\bottomrule
\end{longtable}

\begin{figure}
\centering
\includegraphics{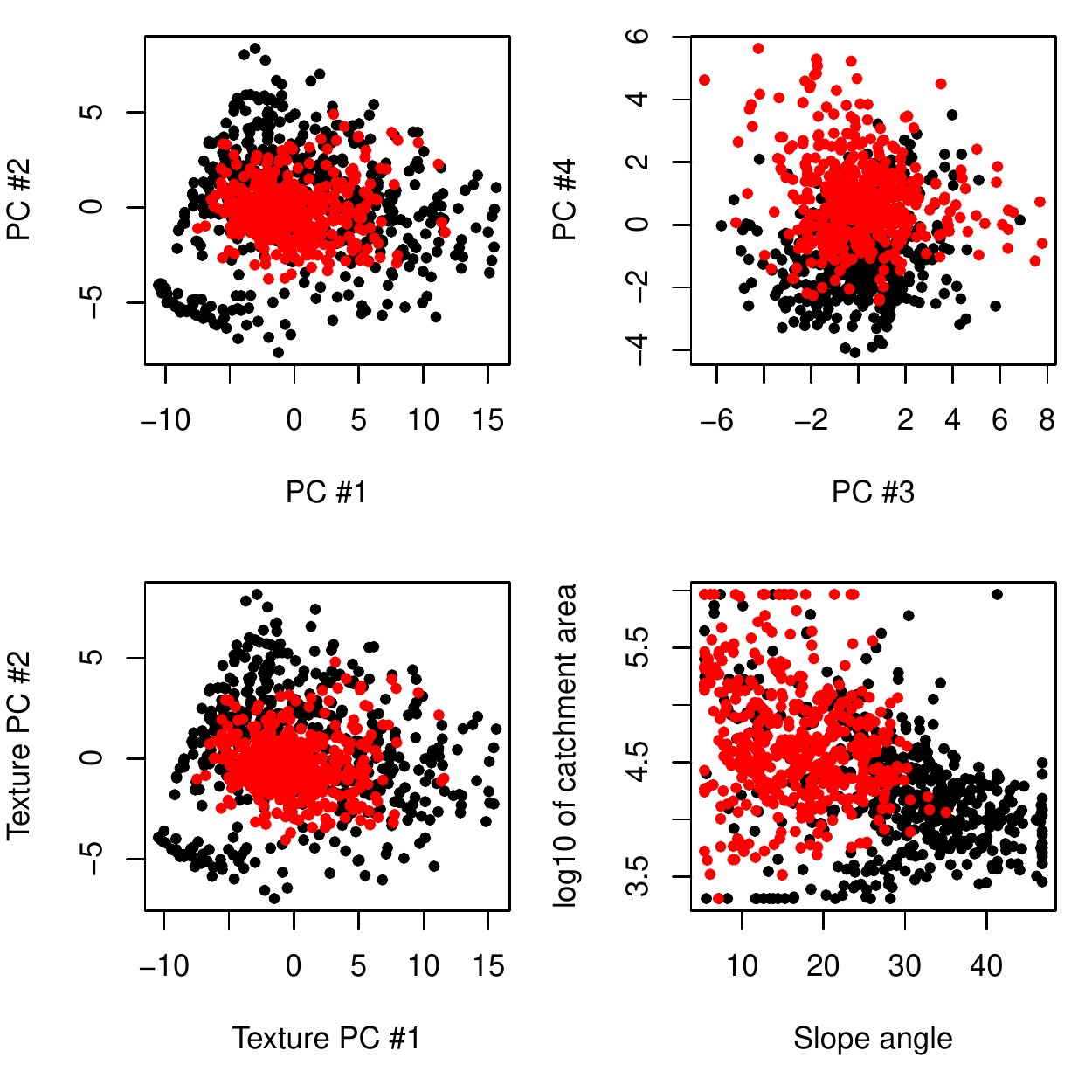}
\caption{\label{fig:featurespace}Feature (sub)space diagrams for the first PCs of the entire feature set.}
\end{figure}

To explore the feature sets, PCAs were performed for the entire set of 46 feature and for the subset of 40 Gabor features (Figure \ref{fig:featurespace}). In the entire feature set, 63.6\% of the variance is concentrated in the first two PCs (first six PCs: 83.7\%). In the more strongly correlated Gabor feature set, in contrast, the first two PCs make up 72.2\% of the variance (first six PCs: 89.5\%). The main PCs turned out to be interpretable by domain experts. PC \#1 of the Gabor feature set (`Gabor1,' in the figures) is basically an overall average of all texture features, meaning that it expresses the overall presence of striped patterns of any characteristics. Gabor PC \#2 represents the contrast between minimum and median aggregated anisotropic Gabor features and the rest; large values are interpreted as incoherent patterns with no distinct, repeated stripe pattern. Gabor PC \#3 expresses differences between large-wavelength range or maximum-aggregated features versus the short-wavelength features, which represents the heterogeneity in the width of stripes, and thus the size of linear surface structures. Very large values correspond to distinct patterns of large amplitude.

A random forest classifier is used for the classification of rock glaciers based on the features introduced above. Its overall accuracy, estimated by spatial cross-validation between the two sub-regions (Brenning 2012), is \(80.1\)\%. Omitting terrain attributes from the feature set has a greater impact on performance than omitting the texture features (Table \ref{tab:accuracies}).

\hypertarget{results}{%
\section{Results}\label{results}}

\begin{figure}
\centering
\includegraphics{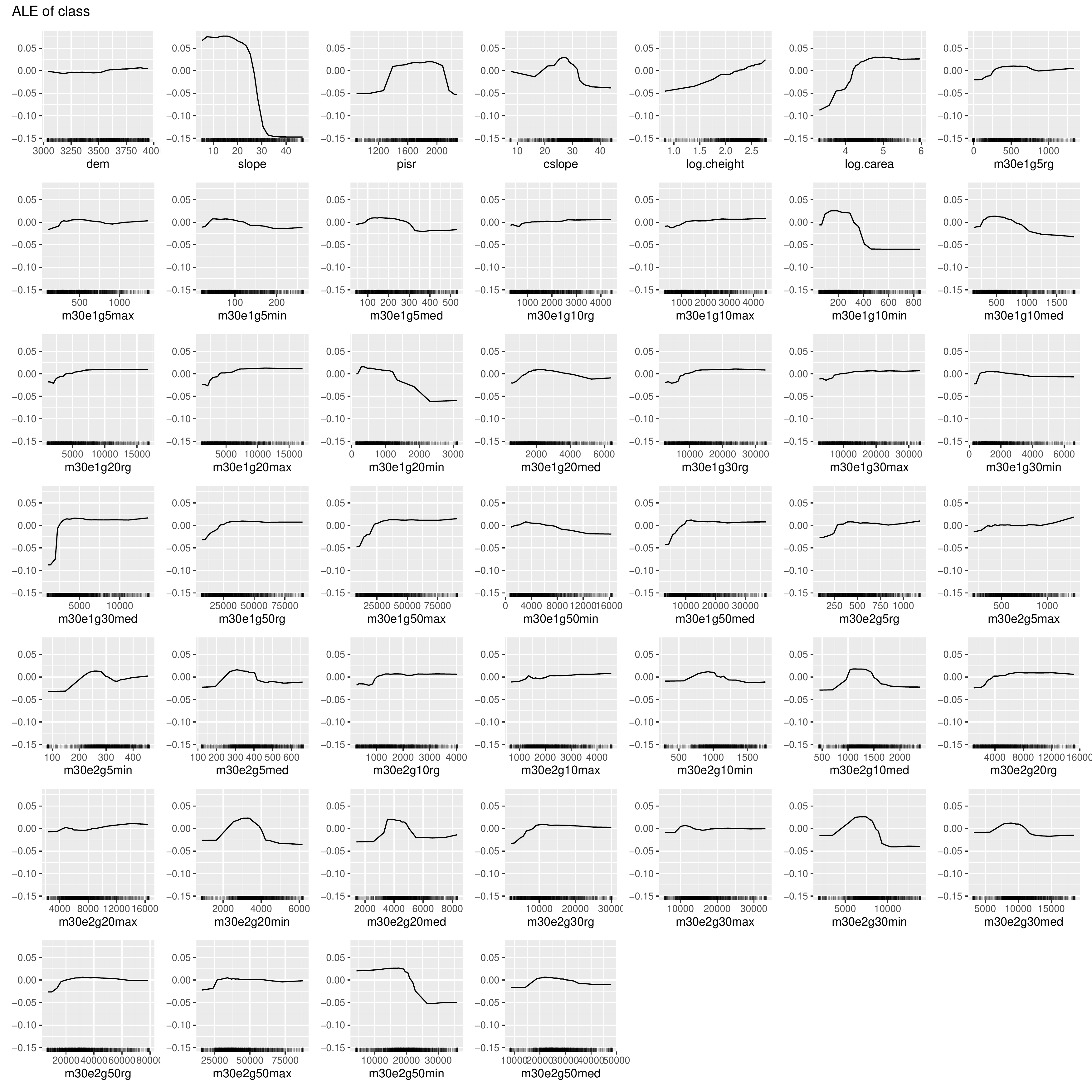}
\caption{\label{fig:simpleeffects}Ordinary ALE plots for all 46 features.}
\end{figure}

With 46 features that are grouped into two semantic feature types (terrain attributes, texture features), it can be challenging to interpret the patterns represented by marginal effects plots (Figure \ref{fig:simpleeffects}). Although there appears to be some consistency in direction among many of the texture features, it is difficult to identify an overall pattern that can be summarized verbally, and it would be unreasonable to present such detailed visual information to a conference audience that is expecting a concise and coherent narrative.

\begin{figure}
\centering
\includegraphics{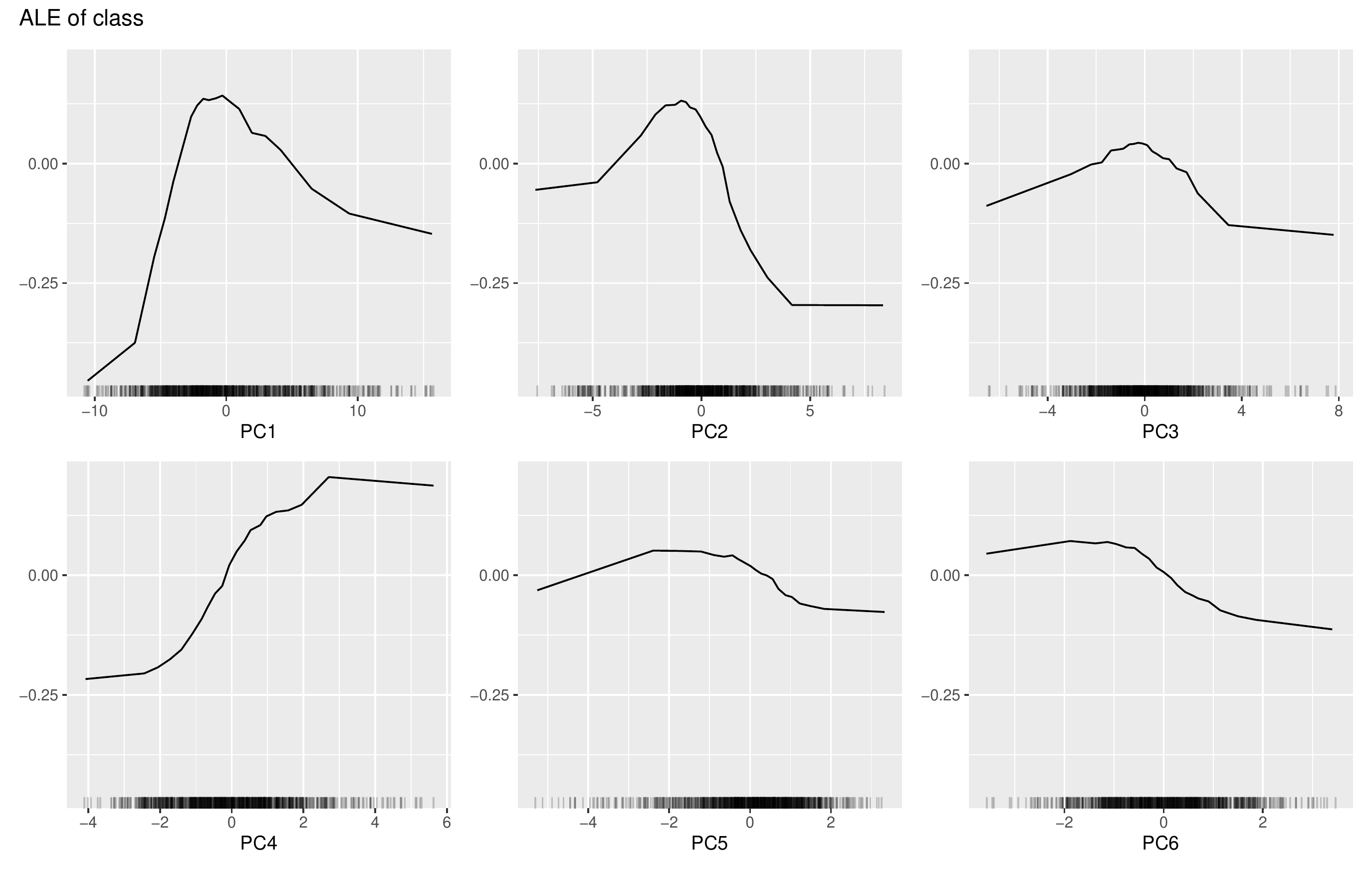}
\caption{\label{fig:simplywarpedeffects}ALE plots along the first six principal axes, applying PCA to the entire feature set.}
\end{figure}

The ALE plots along principal axes distill 71.6 percent of the feature variance into only three plots (Figure \ref{fig:simplywarpedeffects}). Nevertheless, considering the semantic differences and weak correlations between texture features and terrain attributes, it seems unnecessary to combine all features in a joint PCA, which results in PCs with an at least slightly mixed meaning.

\begin{figure}
\centering
\includegraphics{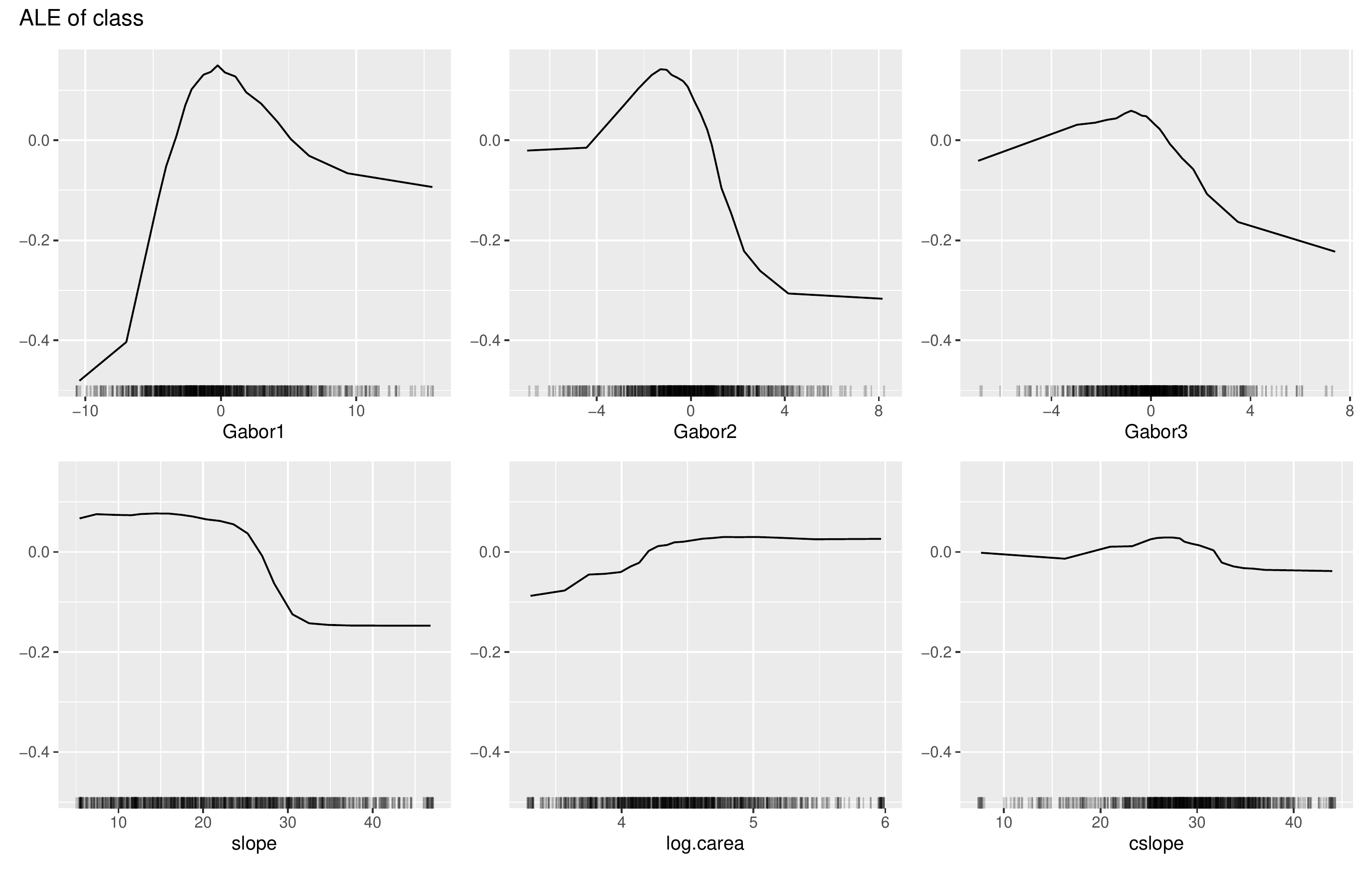}
\caption{\label{fig:strucwarpedeffects}ALE plots along the first principal axes of texture features, and for the most important terrain attributes.}
\end{figure}

The structured PCA approach, in contrast, allows us to explicitly separate the model's representation of effects of texture features and terrain attributes, which is desitable from a domain expert's perspective and statistically justifiables based on the weak correlations between these feature groups. Larger overall texture signals (Gabor PC \#1) are associated with higher predicted rock glacier probability (Figure \ref{fig:strucwarpedeffects}). However, a large contrast between minimum/median anisotropic texture features and the remaining texture features, as expressed by a high Gabor PC \#2 value, is more often associated with an absence of rock glaciers. In other words, the absence of coherently oriented, repeated stripes is not typical of rock glaciers - these may be more typical of non-repeated stripes (e.g., erosion gullies, jagged rock slopes).

\begin{figure}
\centering
\includegraphics{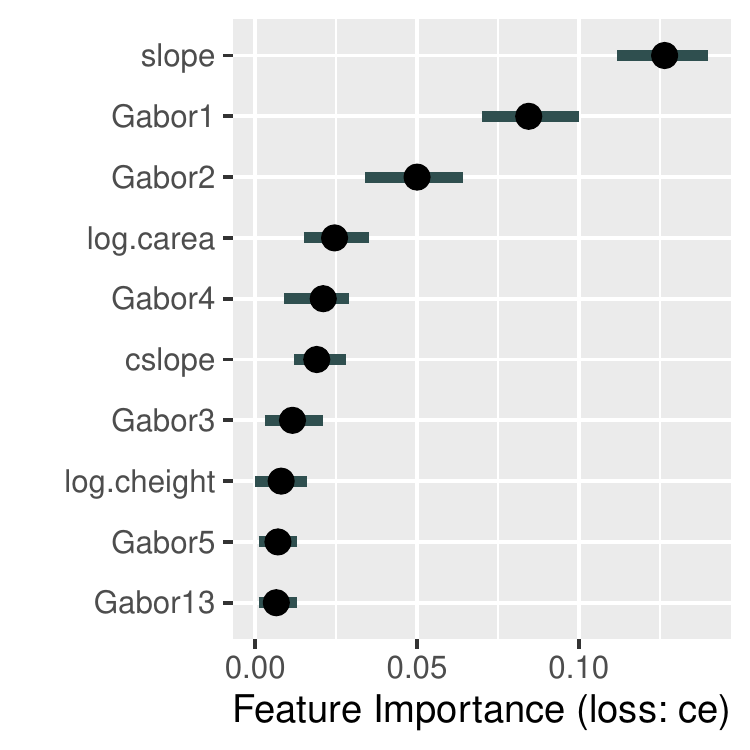}
\caption{\label{fig:featureimportance}Permutation feature importances of the 10 top-ranked texture principal components and terrain attributes.}
\end{figure}

The permutation-based assessment of the importance of texture PCs and terrain attributes shows that subsequent PCs contribute much less to the predictive performance, and that slope angle is the most salient feature overall (Figure \ref{fig:featureimportance}). Clearly, the combined importance of Gabor features as summarized by Gabor PCs \#1 and \#2 provides a more comprehensible summary than an incoherent litany of individual feature importances of strongly correlated features, which should not be permuted independently of each other.

\begin{figure}
\centering
\includegraphics{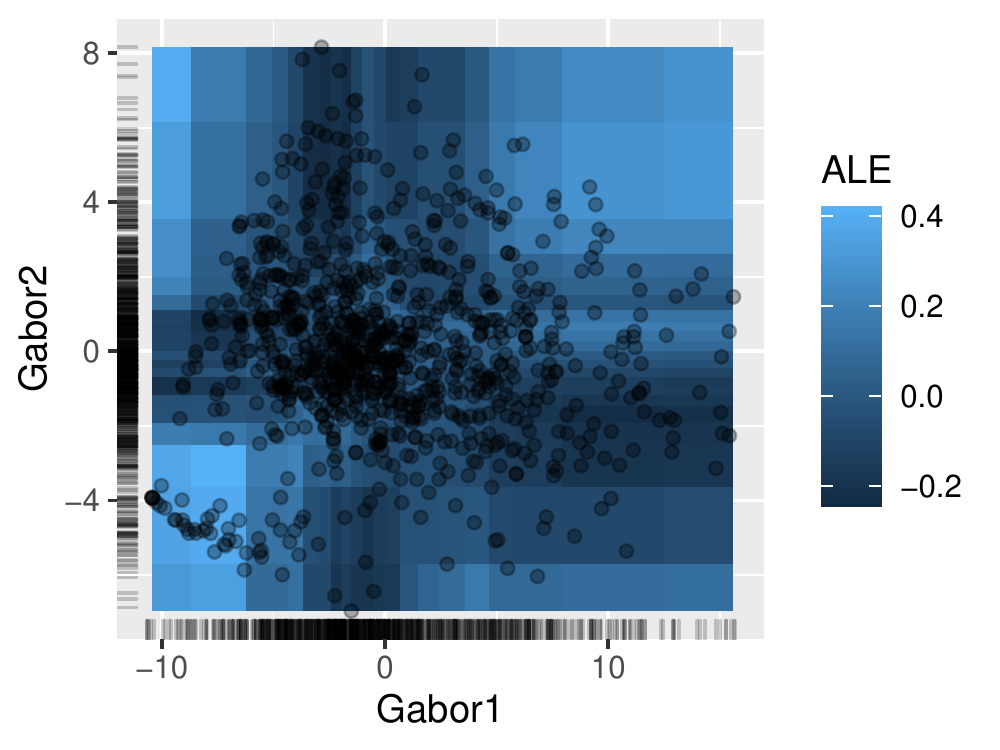}
\caption{\label{fig:strucwarpedeffects2d}Two-dimensional ALE plot with respect to the first and second principal axes of the texture features.}
\end{figure}

\hypertarget{discussion}{%
\section{Discussion}\label{discussion}}

Overall, interpretation plots along the principal axes are capable of distilling complex high-dimensional relationships into low-dimensional summaries, thus providing a tidier, better structured and more focused approach to model interpretation than traditional tools that focus on individual predictors in an ocean of highly correlated features. This behaviour is highly desirable from a domain expert's perspective, and applying it in a structured manner allows the analyst to honour domain knowledge and feature semantics.

Of course fitting the classifier to PCA-transformed data as input features could have provided direct access to ALE plots along principal axes. However, we would want our feature engineering decisions to be directed towards improving predictive performance, and we would therefore prefer not to risk compromising an optimal performance to satisfy our desire to interpret our model. While this is not an issue in the present case study (overall accuracy 0.791 with PC features versus 0.801 with the original predictors), our experience shows that PCA-transformed predictors can worsen the predictive performance. Also, model-agnostic post-hoc analysis tools are precisely meant to be applicable to black-box models that are provided `as is,' without the possibility of altering their input features, in which case the proposed `hands-off' access to transformed perspectives is particularly valuable.

The proposed use of PCA and related linear transformation technique appears to be in contradiction to the use of complex nonlinear machine-learning models. Nevertheless, it could be argued that linear cross-sections of feature space along the original feature axes are no less arbitrary and limiting, considering the often strong correlations with other features that would have to be interpreted simultaneously. From that perspective, principal axes provide a `tidier' perspective and smarter peek into feature space than traditional ALE or partial dependence plots. Linear transformations similar to PCA may further enhance interpretability by offering a more structured or target oriented perspective based on simple components (Rousson and Gasser 2004) or discriminant functions (Cunningham and Ghahramani 2015).

One could even argue that ALE plots can be misleading for highly correlated features as they look at the often tiny contributions of individual features in an isolated way, while the proposed approach focuses on the bigger picture and captures the combined effect of a bundle of features. This also becomes evident in permutation-based feature importance assessments, where individual texture features consistently achieve discouragingly low importances, while the first two PCs of the texture features are ranked very highly. For the specific case of permutation assessments, it has also been proposed to jointly permute groups of features (Molnar 2019); unlike the techniques proposed here, this approach is not transferable to other interpretation tools that are not based on permutations.

Beyond linear transformations, the proposed approach provides a general framework even for nonlinear perspectives on feature space and model functions. In particular, paths proposed in section \ref{following-paths} may well be nonlinear, as e.g.~defined by a physical model that could be used by domain experts to check model plausibility. Also, curvilinear component analyses (CCA) or autoencoders as state-of-the-art multivariate nonlinear transformation methods provide a logical extension of PCA and highlight the link between explainable machine-learning and projection-based visual analytics (Schaefer et al. 2013).

\begin{figure}
\centering
\includegraphics{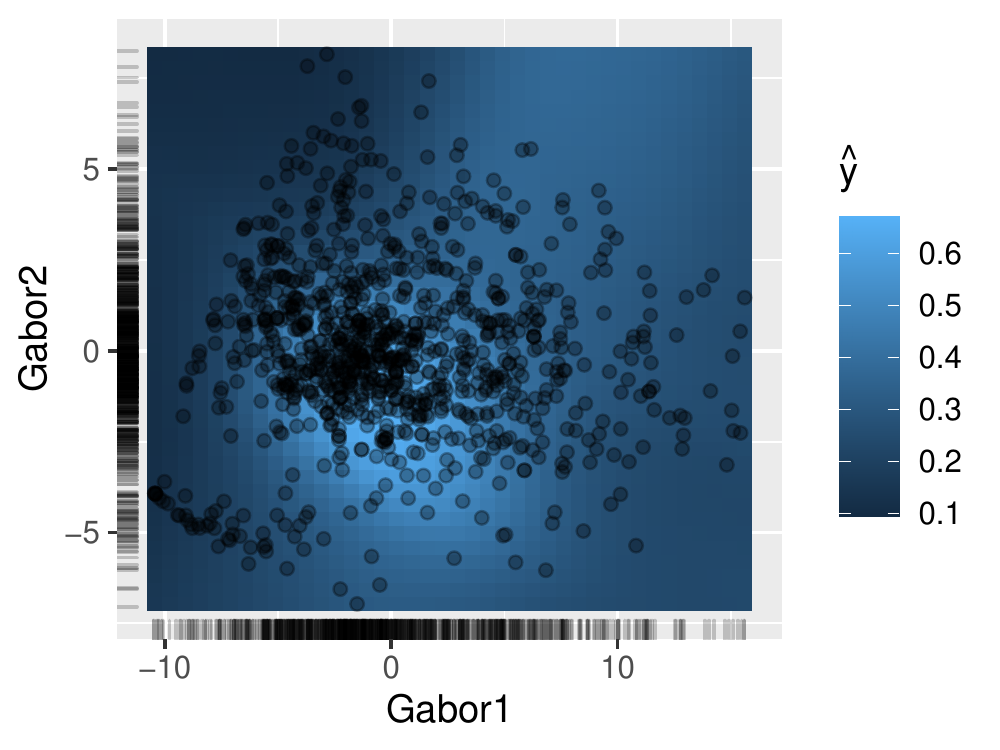}
\caption{\label{fig:strucwarpedeffects2dpdp}Two-dimensional partial dependence plot with respect to the first and second principal axes of the texture features.}
\end{figure}

Finally, due to the orthogonality and thus linear independence of PCs, the more naturally interpretable partial dependence plots become a more viable option for the interpretation of black-box machine-learning models. In original feature space, in contrast, the less intuitive and sometimes rather coarse ALE plots should usually be preferred despite their limitations (compare Figures \ref{fig:strucwarpedeffects2d} and \ref{fig:strucwarpedeffects2dpdp}).

\hypertarget{conclusions}{%
\section{Conclusions}\label{conclusions}}

Despite the inherent limitations of post-hoc machine-learning model interpretation, feature space transformations, and structured PCA transformations in particular, are a powerful tool that allows us to distill complex nonlinear relationships into an even smaller number of univariate plots than previously possible, representing perspectives that are informed by domain knowledge. These transformations provide an intuitive access to feature space, which can be easily wrapped around existing model implementations. Model interpretation through the lens of feature transformation and dimension reduction allows us to peek into the feature space at an oblique angle - a strategy that many of us have have successfully applied when checking if our kids are asleep in their beds, and a much more successful strategy than staring along the walls, i.e.~the original feature axes, especially when these are nearly parallel.

\hypertarget{acknowledgements}{%
\subsection*{Acknowledgements}\label{acknowledgements}}
\addcontentsline{toc}{subsection}{Acknowledgements}

I would like to thank Philipp Lucas (DLR Institute for Data Science) for encouraging discussions on an early version of the manuscript.

\hypertarget{references}{%
\subsection*{References}\label{references}}
\addcontentsline{toc}{subsection}{References}

\hypertarget{refs}{}
\begin{CSLReferences}{1}{0}
\leavevmode\hypertarget{ref-adebayo.kagal.2016.ortho}{}%
Adebayo, J., and L. Kagal. 2016. {``Iterative Orthogonal Feature Projection for Diagnosing Bias in Black-Box Models.''} \url{http://arxiv.org/abs/1611.04967}.

\leavevmode\hypertarget{ref-apley.zhu.2020.ale}{}%
Apley, D. W., and J. Zhu. 2020. {``Visualizing the Effects of Predictor Variables in Black Box Supervised Learning Models.''} \emph{Journal of the Royal Statistical Society: Series B (Statistical Methodology)} 82 (4): 1059--86. \url{https://doi.org/10.1111/rssb.12377}.

\leavevmode\hypertarget{ref-baniecki.biecek.2019.modelstudio}{}%
Baniecki, H., and P. Biecek. 2019. {``{modelStudio}: Interactive Studio with Explanations for {ML} Predictive Models.''} \emph{Journal of Open Source Software} 4 (43): 1798. \url{https://doi.org/10.21105/joss.01798}.

\leavevmode\hypertarget{ref-basille.et.al.2008.habitat}{}%
Basille, M., C. Calenge, E. Marboutin, R. Andersen, and J.-M. Gaillard. 2008. {``Assessing Habitat Selection Using Multivariate Statistics: Some Refinements of the Ecological-Niche Factor Analysis.''} \emph{Ecological Modelling} 211 (1): 233--40. \url{https://doi.org/10.1016/j.ecolmodel.2007.09.006}.

\leavevmode\hypertarget{ref-biecek.2018.dalex}{}%
Biecek, P. 2018. {``DALEX: Explainers for Complex Predictive Models in r.''} \emph{Journal of Machine Learning Research} 19 (84): 1--5. \url{https://jmlr.org/papers/v19/18-416.html}.

\leavevmode\hypertarget{ref-brenning.2012.sperrorest}{}%
Brenning, A. 2012. {``Spatial Cross-Validation and Bootstrap for the Assessment of Prediction Rules in Remote Sensing: The r Package Sperrorest.''} In \emph{2012 IEEE International Geoscience and Remote Sensing Symposium}, 5372--75. \url{https://doi.org/10.1109/IGARSS.2012.6352393}.

\leavevmode\hypertarget{ref-brenning.et.al.2012.gabor}{}%
Brenning, A., S. Long, and P. Fieguth. 2012. {``Detecting Rock Glacier Flow Structures Using {Gabor} Filters and {IKONOS} Imagery.''} \emph{Remote Sensing of Environment} 125: 227--37. \url{https://doi.org/10.1016/j.rse.2012.07.005}.

\leavevmode\hypertarget{ref-clausi.et.al.2000}{}%
Clausi, D. A., and M. E. Jernigan. 2000. {``Designing {Gabor} Filters for Optimal Texture Separability.''} \emph{Pattern Recognition} 33 (11): 1835--49. \url{https://doi.org/10.1016/S0031-3203(99)00181-8}.

\leavevmode\hypertarget{ref-cunningham.ghahramani.2015.jmlr}{}%
Cunningham, J. P., and Z. Ghahramani. 2015. {``Linear Dimensionality Reduction: Survey, Insights, and Generalizations.''} \emph{Journal of Machine Learning Research} 16 (89): 2859--2900. \url{http://jmlr.org/papers/v16/cunningham15a.html}.

\leavevmode\hypertarget{ref-friedman.2001.pdp}{}%
Friedman, J. 2001. {``Greedy Function Approximation: A Gradient Boosting Machine.''} \emph{The Annals of Statistics} 29 (5): 1189--1232.

\leavevmode\hypertarget{ref-greenwell.2017.pdp}{}%
Greenwell, Brandon M. 2017. {``Pdp: An {R} Package for Constructing Partial Dependence Plots.''} \emph{The R Journal} 9 (1): 421--36. \url{https://journal.r-project.org/archive/2017/RJ-2017-016/index.html}.

\leavevmode\hypertarget{ref-hooker.mentch.2019.stop}{}%
Hooker, G., and L. Mentch. 2019. {``Please Stop Permuting Features: An Explanation and Alternatives.''} \url{http://arxiv.org/abs/1905.03151}.

\leavevmode\hypertarget{ref-landgrebe.2002.hyperspectral}{}%
Landgrebe, D. 2002. {``Hyperspectral Image Data Analysis.''} \emph{IEEE Signal Processing Magazine} 19 (1): 17--28. \url{https://doi.org/10.1109/79.974718}.

\leavevmode\hypertarget{ref-molnar.2019.iml.book}{}%
Molnar, C. 2019. \emph{Interpretable Machine Learning: A Guide for Making Black Box Models Explainable}.

\leavevmode\hypertarget{ref-molnar.et.al.2018.iml}{}%
Molnar, C., B. Bischl, and G. Casalicchio. 2018. {``Iml: An {R} Package for Interpretable Machine Learning.''} \emph{JOSS} 3 (26): 786. \url{https://doi.org/10.21105/joss.00786}.

\leavevmode\hypertarget{ref-molnar.et.al.2020.review}{}%
Molnar, C., G. Casalicchio, and B. Bischl. 2020. {``Interpretable Machine Learning -- a Brief History, State-of-the-Art and Challenges.''} \url{http://arxiv.org/abs/2010.09337}.

\leavevmode\hypertarget{ref-molnar.et.al.2020.conditional}{}%
Molnar, C., G. König, B. Bischl, and G. Casalicchio. 2020. {``Model-Agnostic Feature Importance and Effects with Dependent Features -- a Conditional Subgroup Approach.''} \url{http://arxiv.org/abs/2006.04628}.

\leavevmode\hypertarget{ref-molnar.et.al.2020.pitfalls}{}%
Molnar, C., G. König, J. Herbinger, T. Freiesleben, S. Dandl, C. A. Scholbeck, G. Casalicchio, M. Grosse-Wentrup, and B. Bischl. 2020. {``Pitfalls to Avoid When Interpreting Machine Learning Models.''} \url{http://arxiv.org/abs/2007.04131}.

\leavevmode\hypertarget{ref-mountrakis.et.al.2011.svm}{}%
Mountrakis, G., J. Im, and C. Ogole. 2011. {``Support Vector Machines in Remote Sensing: A Review.''} \emph{ISPRS Journal of Photogrammetry and Remote Sensing} 66 (3): 247--59. \url{https://doi.org/10.1016/j.isprsjprs.2010.11.001}.

\leavevmode\hypertarget{ref-murdoch.et.al.2019.iml}{}%
Murdoch, W. J., C. Singh, K. Kumbier, R. Abbasi-Asl, and B. Yu. 2019. {``Definitions, Methods, and Applications in Interpretable Machine Learning.''} \emph{Proceedings of the National Academy of Sciences} 116 (44): 22071--80. \url{https://doi.org/10.1073/pnas.1900654116}.

\leavevmode\hypertarget{ref-pena.brenning.2015.maipo}{}%
Peña, M. A., and A. Brenning. 2015. {``Assessing Fruit-Tree Crop Classification from {Landsat}-8 Time Series for the {Maipo} Valley, {Chile}.''} \emph{Remote Sensing of Environment} 171: 234--44. \url{https://doi.org/10.1016/j.rse.2015.10.029}.

\leavevmode\hypertarget{ref-pena.et.al.2017.spectrotemporal}{}%
Peña, M. A., R. Liao, and A. Brenning. 2017. {``Using Spectrotemporal Indices to Improve the Fruit-Tree Crop Classification Accuracy.''} \emph{ISPRS Journal of Photogrammetry and Remote Sensing} 128: 158--69. \url{https://doi.org/10.1016/j.isprsjprs.2017.03.019}.

\leavevmode\hypertarget{ref-rousson.gasser.2004.sca}{}%
Rousson, V., and T. Gasser. 2004. {``Simple Component Analysis.''} \emph{Applied Statistics} 53 (4): 539--55. \url{https://doi.org/10.1111/j.1467-9876.2004.05359.x}.

\leavevmode\hypertarget{ref-schaefer.et.al.2013.projection}{}%
Schaefer, M., L. Zhang, T. Schreck, A. Tatu, D. A. Keim, J. A. Lee, and M. Verleysen. 2013. {``Improving Projection-Based Data Analysis by Feature Space Transformations.''} In \emph{Visualization and Data Analysis 2013}, edited by Pak Chung Wong, 8654:8654OH--. Proceedings SPIE. SPIE, The Soc. for Imaging Science; Technology. \url{https://doi.org/10.1117/12.2000701}.

\leavevmode\hypertarget{ref-strobl.et.al.2008.conditional}{}%
Strobl, C., A.-L. Boulesteix, T. Kneib, T. Augustin, and A. Zeileis. 2008. {``Conditional Variable Importance for Random Forests.''} \emph{BMC Bioinformatics} 9 (1): 307. \url{https://doi.org/10.1186/1471-2105-9-307}.

\leavevmode\hypertarget{ref-zortea.et.al.2007.highdim}{}%
Zortea, M., V. Haertel, and R. Clarke. 2007. {``Feature Extraction in Remote Sensing High-Dimensional Image Data.''} \emph{IEEE Geoscience and Remote Sensing Letters} 4 (1): 107--11. \url{https://doi.org/10.1109/LGRS.2006.886429}.

\end{CSLReferences}

\end{document}